\documentclass{article}

\usepackage[preprint]{nips_2018}

\usepackage[utf8]{inputenc} 
\usepackage[T1]{fontenc}    
\usepackage{hyperref}       
\usepackage{url}            
\usepackage{booktabs}       
\usepackage{amsfonts}       
\usepackage{nicefrac}       
\usepackage{microtype}      
\usepackage{algorithm}
\usepackage[small]{caption}
\usepackage{amssymb}
\usepackage{amsmath}
\usepackage{verbatim}
\usepackage[noend]{algpseudocode}
\usepackage{graphicx}
\usepackage{subcaption}
\usepackage{booktabs}
\usepackage{todonotes}

\makeatletter
\def\BState{\State\hskip-\ALG@thistlm}
\makeatother

\title{Distributional Negative Sampling for Knowledge Base Completion}

\author{
  Sarthak Dash \\
  IBM Research AI\\
  \texttt{sdash@us.ibm.com} \\
  \And
  Alfio Gliozzo \\
  IBM Research AI \\
  \texttt{gliozzo@us.ibm.com} \\
}

\begin{document}

\maketitle

\begin{abstract}

State-of-the-art approaches for Knowledge Base Completion (KBC) exploit deep neural networks trained with both false and true assertions: positive assertions are explicitly taken from the knowledge base, whereas negative ones are generated by random sampling of entities. In this paper, we argue that random sampling is not a good training strategy since it is highly likely to generate a huge number of nonsensical assertions during training, which does not provide relevant training signal to the system. Hence, it slows down the learning process and decreases accuracy. To address this issue, we propose an alternative approach called Distributional Negative Sampling that generates meaningful negative examples which are highly likely to be false. Our approach achieves a significant improvement in \textit{Mean Reciprocal Rank values} amongst two different KBC algorithms in three standard academic benchmarks. 

\end{abstract}

\section{Introduction} \label{introduction}

Knowledge Base Completion (KBC) systems leverage existing knowledge from within an incomplete input KB in order to validate the truth value of unknown assertions, so as to augment an existing KB with additional assertions. A sub-task within KBC called \textit{link prediction} is defined as follows: given the predicate and either subject or object entity of an assertion, the system should correctly predict the missing entity. 
For every possible entity in the KB an assertion is generated, and the system provides an assessment of its validity which it then uses to generate a ranked list of all possible candidates.

One of the popular approaches to train models for KBC leverage the assertions in the original KB in order to provide positive examples and use \textit{Random Negative Sampling} (RNS) to generate negatives. RNS for link prediction consists of selecting a triple from the KB and substituting one of its entity (e.g. the object) with a random entity from the KB. For example, a positive assertion \textit{PlaysInMovie}(Elijah Wood, Lord Of The Rings) might generate \textit{PlaysInMovie}(Elijah Wood, DVR) or \textit{PlaysInMovie}(Elijah Wood, Star Trek) as negative examples. 
However, this approach is far from being the optimal choice for training KBC systems, which are supposed to distinguish between true and false assertions. In fact, if one looks in depth into the meaning of the two different negative assertions generated by RNS in the example above, there is a very big difference: the first one does not make sense, while the second is actually meaningful, but false. 

In this work, we introduce Distributional Negative Sampling (DNS), an alternative approach for RNS that can be applied to train deep learning based KBC systems. DNS is aimed at generating meaningful negative examples which are highly likely to be false, preventing generation of nonsensical negatives. DNS uses distributional similarity to drive the negative sampling process. Specifically during training, given a positive triple, DNS generates negative examples by replacing that entity with other entities that are similar to it. In the example above, the vector for \textit{Lord Of The Rings}, as estimated during the KBC learning process at that state is compared against all other entities in the KG; those having higher similarity are more likely to be chosen as negative examples as opposed to random ones. As an example, the representation for \textit{StarTrek} is highly likely to be similar to \textit{Lord Of The Rings}, as opposed to \textit{DVR}, and therefore the assertion \textit{PlaysInMovie}(Elijah Wood, Star Trek) is more likely to be selected as a negative sample. This creates a repository of meaningful (and mostly false) statements, that we use as negative examples for that statement.

Our intuition is that the entity vectors trained by KBC deep nets represent the distributional properties of entities in the original KB. Therefore, cosine similarity between those vectors can be used as a proxy for their distributional similarity. Since entity vectors are updated dynamically by the KBC algorithm, the behavior of DNS improves at every epoch, sparking a virtuous circle that reduces substantially the number of epochs needed for  while significantly improving accuracy.   

We test DNS on \emph{two} different KBC architectures, showing consistent and significant improvement in \emph{five} out of \emph{six} settings - \emph{two} architectures across three different evaluation benchmarks. DNS is the main contribution of this paper.

The rest of the paper is structured as follows. In section \ref{rel_work}, we describe some of the common KBC approaches, focusing on the deep learning models and their negative sampling strategies. DNS is described in Section \ref{DNS}. In section \ref{experiments} we describe our evaluation benchmarks, discussing the evaluation results. We also provide an analysis of the algorithm in section \ref{Analysis}. Finally, Section \ref{conclusions} concludes the paper opening some interesting research direction for the future work.

\section{Related Work} \label{rel_work}
KBC is a very active field of research and state of the art approaches for this task are almost entirely deep learning based and make use of RNS. \citet{bordes2013translating} define an algorithm \emph{TransE} wherein the relation vector translates the subject representation onto the object representation. \citet{wang2014knowledge} define an algorithm \emph{TransH} that behaves like \emph{TransE} with a minor adjustment that the translation happens on relation specific hyper-planes upon which the entity vectors are projected. Compositional models using the tensor product such as \emph{RESCAL} \citet{nickel2011three} and \emph{Neural Tensor Network} \citet{socher2013reasoning} employ higher order tensors to represent entities and relations. \citet{nickel2016holographic} employs correlation operator between entity representations and a dot product with the relation vector, to represent interactions within the triples for their HolE system. \citet{trouillon2016complex} represent each real valued embeddings into their corresponding complex variants, and performs Hermitian dot product operation - the argument being the fact that this operation models both symmetric and antisymmetric relations. More complex models have been developed that use the path and content information in the KGs, for instance see \citet{lin2015modeling,toutanova2015observed}.
\par

Just an handful of work in the KBC proposes alternatives to RNS. \citet{xie2017interpretable} introduce a \textit{domain sampling approach}, where for each relation $r$ a probability $p_r$ (based on the training data) is calculated. The negative sample is sampled from within the domain with probability $p_r$ and with probability $1-p_r$ it is sampled from the set of all entities. \citet{cai2017kbgan} define a GAN, an adversarial learning framework that uses the generator network to provide negative samples to the discriminator. \citet{kotnis2017analysis} proposes an approach to negative sampling based on using additional external knowledge about the type of the entities in the KB. 
\par
This approach consists of two steps, first a KBC model is trained using Typed-Sampling i.e. corrupted entities belonging to the same type as the replacement entity are used to construct negative samples. In the second step,
a standard KBC model is trained which uses nearest neighbor sampling (on the frozen model in the first step) to generate negative samples. Since, such a system requires additional knowledge, this approach cannot be applied to KB where such type information is sparse or not available, making its applicability to real word scenarios unfeasible.

DNS approach is different from above work in the following fashion. First, DNS does not use type information and any other additional assumption. Second, DNS uses a stochastic approach to sample negatives as opposed to a deterministic nearest neighbor approach.

Note that, DNS reuses the embedding updated within the same KBC network it is has been trained from, generating a positive feedback loop that ultimately results in faster convergence (shown empirically) as measured by the number of epochs.

\par 
\section{Distributional Negative Sampling} \label{DNS}
Let us define the problem setting in terms of notations. Denote the knowledge base consisting of a list of true triples as, \[KB=\{(h, r, t)_i\}_{i=1}^{n}\] where $(h, r, t)_i$ denotes the $i^{th}$ fact present in the knowledge base comprising of head entity, relation and tail entity respectively. Let $n$ denote the total number of triples present in the knowledge base, and denote the set of entities and relations in the KB by $\mathcal{E}$ and $\mathcal{R}$. 
The problem statement for KB Completion is to learn representations of entities and relations within $\mathcal{E}$ and $\mathcal{R}$ that best predict the triples present in the $KB$. Most deep learning architectures for KBC exploit fixed dimensional tensors to represent entities and relations; and differ by the way they combine these tensors to score a given triple - this difference yields different architectures. 

\par 

Training of these models for KBC requires generation of negative samples, which are created under the Local Closed World Assumption (LCWA). Leveraging this assumption, one randomly generates negative examples by corrupting existing triples from the $KB$. We call this set $\mathcal{D_-}$, as opposed to the set  $\mathcal{D_+}$ which contains triples from original KB.

 A commonly used loss function for training (follows from LCWA) is the pairwise ranking loss, 
\begin{equation} \label{eq:1}
  min_\Theta \sum\limits_{i \in \mathcal{D_+}}\sum\limits_{j \in \mathcal{D_-}} max(0, \gamma+f_j-f_i)
\end{equation}
where $\gamma > 0$ specifies the width of the margin \citet{bordes2011learning}. 

\par 

In practice, a fixed number C (a hyper-parameter) of negative samples are generated at random (hereby referred to as Random Negative Sampling(RNS)) in order to create a skewed distribution with a small yet fixed negatives-to-positives ratio. The negative samples are generated by either replacing the head (or tail) entity of a given triple by a random entity from $\mathcal{E}$. This approach is then repeated for all triples in $KB$, followed by calculating the loss function in Equation \ref{eq:1} which is subsequently optimized during training\footnote{For a quick review of machine learning on knowledge graphs that use pairwise ranking loss see also \citet{nickel2016review}.}. 

\par 

\begin{figure*}[ht]
   \centering
   \includegraphics[width=4.0in]{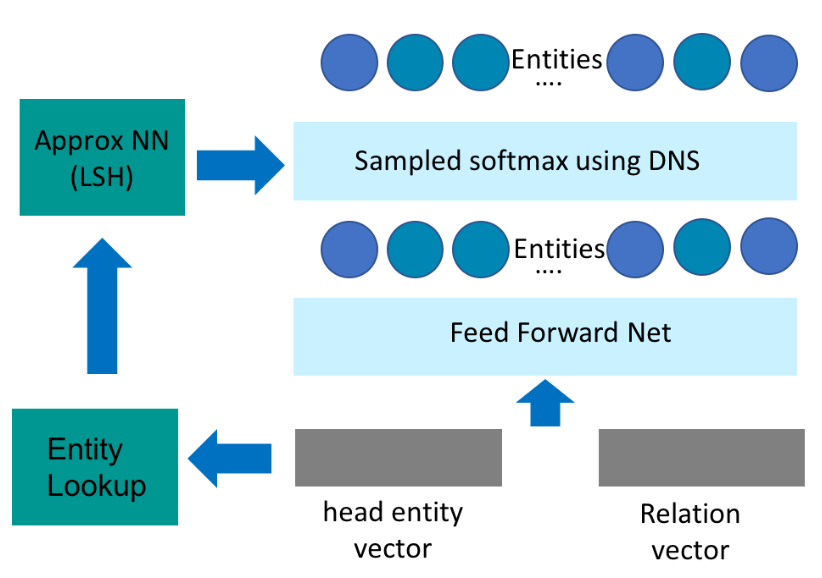}
   \caption{DNS Architecture using an off-the-shelf KBC model.}
   \label{DNSArchitecture}
\end{figure*}

DNS on the other hand, has been designed to provide a solution to the problem of generating nonsensical training examples. DNS approach relies on the idea that meaningful assertion can be automatically generated by replacing entities in a given triple with other entities belonging to the same type. Note that, two entities $e_1$ and $e_2$ tend to have the same type if they share many relations, i.e. they are distributionally similar\footnote{This approach in inspired by the \textit{distributional hypothesis} in computational linguistics \citet{harris1954distributional}.}. Inspired by this idea, we propose the DNS block-diagram in Figure \ref{DNSArchitecture}.

Our intuition is that the entity vectors trained by KBC deep networks represent well the distributional properties of entities in the original KB. Therefore, cosine similarity between those vectors can be used as a proxy for the semantic relatedness. This enables us to define the DNS algorithm as described by Algorithm \ref{tab:neg_sample} below. 

\begin{algorithm}
\caption{Distributional Negative Sampling}\label{tab:neg_sample}
\textbf{Input} Training set S = \{(\textit{h,r,t})\}, entity and relation sets $\mathcal{E}$ and $\mathcal{R}$, batch size b. \\ 
\textbf{Returns} Given a batch B = \{$(\textit{$h$,$r$,$t$})_k$\}, $1\leq k \leq b$,
return corrupted triples \textit{P} for each triple in the batch.
\begin{algorithmic}[1]
\BState \emph{$P = \phi$}
\For{\text{each triple $\psi_j$ = $(h,r,t)$ } $\in$ batch B}
\State $N = \phi$ \Comment{ Stores negative triples for triple $\psi_j$}
\State $M \gets \text{cosine-sim($t$, $\mathcal{E}$)}$ \Comment{Use bernoulli sampling to conclude that \textit{t} should be corrupted.}
\For{\text{(i, m) $\in$ enumerate(M)}} \Comment{python style syntax. M: shape $\vert\mathcal{E}\vert$}
\If {$(h,r,m) \in S$} 
\State continue  \Comment{Do not consider samples that are known to be true}
\EndIf
\State $p_{accept} \gets max(0, m)$
\State $p_{reject} \gets 1-p_{accept}$
\State With probability $p_{accept}$ choose entity i.
\If {\text{entity i is chosen}}
\State $N \gets N\cup (h,r,\mathcal{E}[i])$ \Comment{$\mathcal{E}$[i] denotes entity i}
\EndIf
\EndFor
\State \text{Append (j, N) to P} \Comment{j: index of triple $\psi$ in step 2}
\EndFor
\State \text{return }$P$
\end{algorithmic}
\end{algorithm}

\par 
Let us see in detail what's happening in Algorithm [\ref{tab:neg_sample}] below. Let P denote all negative triples constructed per epoch. Steps 3-13 describes the algorithm for each training instance. For each triple $\psi_j$, we use bernoulli sampling (as introduced by \citet{wang2014knowledge}) to determine whether to corrupt the tail or the head entity. Without loss of generality, let us assume that the tail entity $t$ is to be corrupted. Cosine similarity is used to compute similarity values between $t$ and all other entities present in set $\mathcal{E}$\footnote{Note that we have used cosine-similarity as a measure of distributional similarity in Algorithm [\ref{tab:neg_sample}]. We tried other distance metrics such euclidean and manhattan distances, but found that they performed no better.}. If the generated negative sample actually exists in the training set then it is ignored (Steps 6,7) else, we make use of its associated similarity score $m$ in order to convert it into relevant probabilities in steps 8, 9. The intuition is that entities having representations orthogonal (or worse) to the entity in question are hardly similar, and therefore will not generate a meaningful assertion. Now, using the probabilities defined above, we choose whether to include the corresponding entity within our list of negative samples $N$ or not.
\par
Once N is constructed (pertaining to the loop between steps 5-12), this set consists of negative samples corresponding to input triple $\psi_j$. We then append $(j, N)$ i.e. all the corrupted instances for $\psi_j$ to the list P, and repeat this process for all the triples in the training data. Once completed, the list P contains all negative samples generated via DNS for the training data.

A few points to note on the above Algorithm [\ref{tab:neg_sample}]. Firstly, because of steps 8-10, there exists a non-zero albeit small probability of choosing a totally unrelated entity as a negative sample. When repeated over many training epochs, this allows the model to effectively explore the space of negatives, by focusing highly on meaningful negatives and less (non zero) on nonsensical assertions, as opposed to deterministically choosing top k nearest neighbors proposed by \citet{kotnis2017analysis}.

\par 
Thirdly, unlike other KBC algorithms which require the number of negative samples as a hyper parameter (one that needs to be optimized by cross-validation), our Algorithm [\ref{tab:neg_sample}] does not need this parameter to explicitly specified. It automatically figures it out based on sampling step 10, using the accept/reject probabilities defined in steps 8,9 respectively. Thus in this aspect, the DNS variant is more generic than its RNS counterpart.

\par 

Implementation wise, in real-world situations that involve enormous number of entities, computing cosine-similarity in batches during training (step 4 in Algorithm [\ref{tab:neg_sample}]) may not be possible due to out-of-memory issues. To alleviate this, we can make use of Annoy\footnote{https://github.com/spotify/annoy}, an open source library that implements Approximate Nearest Neighbor functionality using Locality-Sensitive Hashing(LSH).

\section{Evaluation} \label{experiments}

In this section we provide an extensive evaluation of DNS and RNS on \textit{link prediction} tasks for three data sets widely used in KBC literature. The \textit{link prediction} task is defined as the task of predicting the correct object (or subject) entity given the subject (or object) entity and the relation parameter. Link prediction task is usually defined by splitting a gold standard KB into two subgraphs, used for training and test.  For each triple in the test the KBC model is queried, this returns a ranked list of entities as an output. This ranked list is then used for evaluation purposes.

We evaluated our system in three different standard link prediction benchmarks, derived from Freebase \citet{bollacker2008freebase} and WordNet \citet{miller1995wordnet}. For the Freebase KB, we use FB15K introduced in \citet{bordes2013translating}, and a more challenging dataset Fb15k-237 introduced in \citet{toutanova2015observed}. The latter is obtained by removing near-duplicate and inverse relations from FB15K. For the WordNet KB, we use the dataset WN18RR introduced by \citet{dettmers2017convolutional}, which removes reversing relations and increases the difficulty of reasoning. Compared to an older WN18 dataset, this dataset WN18RR is much more challenging and therefore serves as a better dataset compared to its older variant. The evaluation metrics that we use are \textit{filtered} Mean Reciprocal Rank (MRR), Hits@10 and Hits@1. The dataset statistics are summarized in Table \ref{tab:dataset}.
\begin{table*}[ht]
    \center
    \caption{Knowledge Base completion datasets statistics} \label{tab:dataset}
    \begin{tabular}{ cccc } 
 \toprule
 Dataset & FB15K & Fb15k-237 & WN18RR \\ \midrule  
 \# Train & 483,152 & 272,115 & 86,835 \\ \hline 
 \# Valid & 50,000 & 17,535 & 3,034 \\ \hline 
 \# Test & 59,071 & 20,466 & 3,134 \\ \hline 
 \# Entities & 14,951 & 14,951 & 40,943 \\ \hline 
 \# Relations & 1,357 & 237 & 11 \\  
 \bottomrule
\end{tabular}
\end{table*}

We compare the performances of DNS and RNS when applied to two different KBC algorithms: TransE \citet{bordes2013translating} and RESCAL \citet{nickel2011three}. These algorithms are easy to visualize in terms of vector operations. The DNS negatives thus generated are used directly to calculate the pairwise ranking loss (as described in equation \ref{eq:1}). 

The hyperparameters for all the experiments were fine-tuned (using grid search) based on the validation set, and \textit{Adam} optimizer \citet{kingma2014adam} was used with default hyper-parameter settings: $\beta_1=0.9, \beta_2=0.999, \epsilon=1e$\textsuperscript{-8}. For FB and WN datasets, RESCAL used an embedding dimension of 200, whereas TransE algorithm used an embedding dimension of 100. The margin parameter for TransE model was finally set to 10.0, and for RESCAL model it equalled 5.0. 
Bernoulli Sampling introduced by \citet{wang2014knowledge} was used to determine whether to corrupt the head or the tail entity. The maximum number of epochs was capped at 1000 for TransE/RESCAL models for FB and WN datasets. In addition, early-stopping criterion was considered to be the Filtered MRR on the validation set, and the threshold was set to be 20 epochs.

Table \ref{tab:results} provides comparable evaluation of RNS and DNS across the three different KBC benchmarks.

\begin{table*}[ht]
\center
    \caption{Knowledge Base completion performance comparison. H@n denotes Hits@n. We report \textit{filtered} Mean Reciprocal Rank (MRR), \textit{filtered} Hits@10 and \textit{filtered} Hits@1 numbers in this table. All the values expressed here are in percentages. Results marked with \dag are produced by running Fast-TransX (\citet{lin2015learning}) with its default parameters. RNS results for TransE and RESCAL are borrowed from \citet{nickel2016holographic}.}
\label{tab:results}
\footnotesize 
\begin{tabular}{cccccccccc}
    \toprule
    & \multicolumn{3}{c}{FB15k} & \multicolumn{3}{c}{Fb15k-237} & \multicolumn{3}{c}{WN18RR}\\ 
    \cmidrule(r){2-4} \cmidrule(r){5-7} \cmidrule(r){8-10}
    Algorithm & MRR & H@10 & H@1 & MRR & H@10 & H@1 & MRR & H@10 & H@1 \\ \midrule 
    TransE(RNS) & \textbf{46.3} & \textbf{74.9} & 29.7 & 25 & 42.8 & 16.9 & - & $\text{43.2}^{\dag}$ & - \\ 
    TransE(DNS) & 43.0 & 63.9 & \textbf{31.1} & \textbf{29.2} & \textbf{45.7} & \textbf{20.9} & \textbf{18.4} & \textbf{44.4} & \textbf{4.3} \\ \hline
    RESCAL(RNS) & 35.4 & 58.7 & 23.5 & 22.6 & 34.4 & 16.3 & 39.9 & 42.1 & 38.6 \\
    RESCAL(DNS) & \textbf{41.2} & \textbf{62.7} & \textbf{29.6} & \textbf{27.5} & \textbf{44.1} & \textbf{19.2} & \textbf{42.8} & \textbf{44.1} & \textbf{42.1} \\  
    \bottomrule
\end{tabular}
\end{table*}

DNS outperforms RNS on all the considered KBC algorithms and across all benchmarks and evaluation metrics\footnote{The only exception to this statement are the TransE filtered MRR and filtered Hits@10 results on FB15k which further warrants a detailed investigation.}. Remarkably, we obtained very good improvement in accuracy (Hits@1), which is arguably the most important metric for real applications.

\section{Analysis} \label{Analysis}

In this section we provide an in depth analysis of the properties of DNS. First, we show that the learned vectors for the entities in the KG represent their meaning nicely from a distributional perspective, thus providing a qualitative analysis of the generated negative samples. Secondly, we show that DNS converges in a lower number of epochs compared to RNS, and we provide an argumentative analysis of why it happens.  

\begin{table*}[ht!]
\center
\caption{An example of negative samples generated by DNS for the entity \textbf{DVD} at various epochs.}
\begin{tabular}{cccc}
    \toprule
    EPOCH 1 & EPOCH 5 & EPOCH 10 & EPOCH 90  \\ \midrule
  English(0.40) & Blu-ray Disc(0.56) & Blu-ray Disc(0.55) & Blu-ray Disc(0.41)  \\ 
  US Dollar(0.37) & VHS(0.50) & VHS(0.53) & VHS(0.34)  \\ 
  Blu-ray Disc(0.36) & video(0.33) & video(0.36) & video(0.27)  \\ 
  executive producer(0.29) & Silent Hill(0.25) & English(0.23) & television(0.18)  \\ 
  French(0.28) & Luther(0.23) & television(0.23) & Kid Rock(0.14)  \\ 
  \bottomrule 
\end{tabular}
\label{tab:our-sampl-approach-example}
\end{table*}

We start from the observation that entity embeddings are learned during training of the KBC system, which is in turn trained using negative examples provided by DNS that rank them according to their similarity with the substituted entity. This creates a reinforcing cycle that in our experience seems to drive the algorithm toward faster convergence. 

\begin{figure}
%
\begin{subfigure}[h]{0.45\linewidth}
\includegraphics[width=\linewidth]{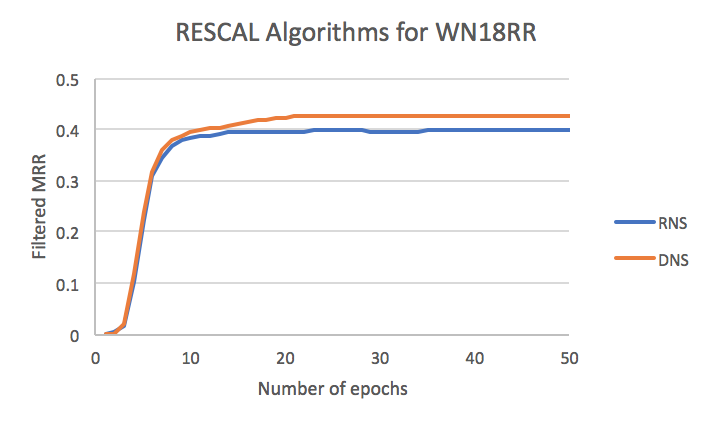}  
\caption{}
\label{fig:RESCAL_WN18RR}
\end{subfigure}
\quad
\begin{subfigure}[h]{0.45\linewidth}
\includegraphics[width=\linewidth]{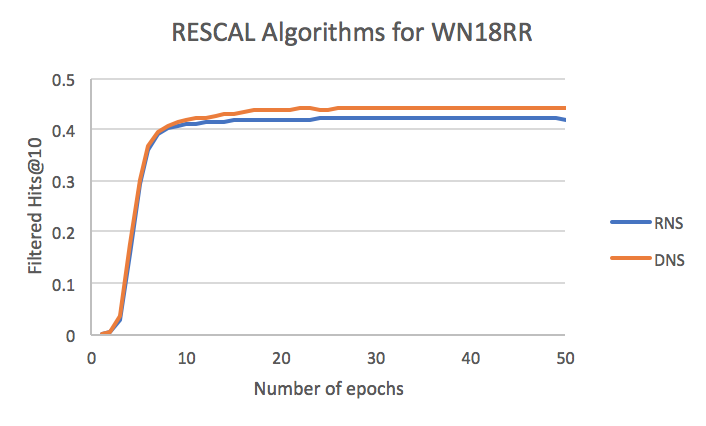}  
\caption{}
\label{fig:RESCAL_WN18RR_H10}
\end{subfigure}

\caption{The left column represents Filtered MRR vs. Number of epochs, whereas the right column represents Filtered Hits@10 vs. Number of epochs. All these figures are plotted against the test data for first 50 epochs only.}
\label{fig:trainingEfficiency}
\end{figure}


Table \ref{tab:our-sampl-approach-example} shows the nearest-neighbor results for the query entity \textit{DVD} after Epochs 1, 5, 10, 90 for the \emph{TransE(DNS)} algorithm on Fb15k-237 dataset. From this table, it is clear that as the training progresses, entities with similar types tend to become closer (via cosine similarity). This enables our DNS algorithm to provide meaningful false statements instead of nonsensical ones. As an example, given the triple \textit{film-distribution-medium(The Day After Tomorrow, DVD)}, RNS and DNS both at Epoch 1, are highly likely to generate nonsensical negatives like \textit{film-distribution-medium(The Day After Tomorrow, US Dollar)}.

However, after a few epochs, things change drastically. At epoch 90 for the same assertion \textit{film-distribution-medium(The Day After Tomorrow, DVD)}, the odds of choosing \textit{VHS} as a negative object by DNS  is 1.05 x $10^{-4}$. Note that, this result is obtained by taking a softmax over the similarity scores computed between the query entity \textit{DVD} and all other entities in Fb15k-237 dataset, and then performing a look-up for the entity \textit{VHS}. Compared to the above, the odds of choosing \textit{VHS} as a negative entity via RNS equals 1/14,952 $\approx 6.6 \text{x} 10^{-5}$. Thus, DNS is twice as likely to generate a meaningful assertion such as \textit{film-distribution-medium(The Day After Tomorrow, VHS)} compared to RNS. Note that as already established before, as per LCWA this assertion is counted as \textit{less positive}, as it's unseen in the training data. 

In summary, by having meaningful positive/negative training facts the system can better classify the validity of unknown assertions. From a machine learning perspective, this way of selecting negative samples provides a more efficient way to train the system.

This is because since RNS is highly likely to generate a nonsensical negative sample, the score of such negatives is highly likely to satisfy the margin in the overall hinge loss function $\mathcal{L}$ (from equation \ref{eq:1}). Comparing this to DNS, during latter stages of training, since the generated corrupted entity is closer to the actual entity (assuming unit normalized vectors), the overall score for such a negative sample is less likely to satisfy the margin, and hence contribute to a non-zero loss and non-zero gradients. 

\par 



We support the above argument for improved efficiency, via empirical fashion as illustrated in Figure \ref{fig:trainingEfficiency}, where both filtered MRR and filtered Hits@10 are reported as a function of the number of epochs (restricted to first 50 epochs) for RESCAL algorithm on WN18RR dataset. In these figures, we see that both MRR and Hits@10 values for Distributional Negative Sampling(DNS) grows more rapidly as compared to Random Negative Sampling(RNS). For example in figure \ref{fig:RESCAL_WN18RR}, both DNS performs slight better compared to RNS until epoch 10 (RNS:0.382 DNS:0.394) after that RNS saturates at 0.4 whereas filtered MRR for DNS rises till 0.43.


\section{Conclusion and Future Work} \label{conclusions}

\par

In this paper, we propose an alternative to Random Negative Sampling (RNS) for Knowledge Base Completion (KBC), namely Distributional Negative Sampling (DNS), that generates meaningful negative examples which are highly likely to be false. We argue that RNS is highly likely to generate nonsensical assertions and we demonstrate how DNS solves this problem in a principled fashion. DNS consistently improves almost all the evaluation metrics over three widely known benchmarks for the two considered algorithms. 

Although in this paper we focus specifically on KBC task, we believe that DNS is a general approach that can be used across many different tasks (that involve generating corrupted units as negative instances) where until now RNS has been used. A typical example of such a task is relation extraction. 

We are interested in building deep learning-based reasoning systems, and we believe that DNS is just the first step in this direction as it plays a pivotal role in building better KBC models for common sense reasoning i.e. the ability to realize that some facts hold purely due to other existing relations. For KBC task, DNS provides a way for efficient training since it generates plausible false statements in a very efficient and mathematically principled manner without any further assumption. 

\par

\appendix

\bibliographystyle{named}
\bibliography{nips_2018}

\end{document}